%% file: main.tex
\documentclass{myThesisTemp}

\usepackage{booktabs}
\usepackage[outdir=./]{epstopdf}
\title{Plug-and-Play Multi-Concept Adaptive Blending for High-Fidelity Text-to-Image Synthesis}{고품질 텍스트-투-이미지 생성을 위한 플러그-앤-플레이 다중 개념 적응적 융합 기법}

\author[korean]{우 영 범}{}{}
\author[english]{Woo}{Young-Beom}{}

\advisor{이 성 환}{Seong-Whan Lee}

\department{
Artificial Intelligence
}{
인 공 지 능 학 과
}

\committee[1]{Seong-Whan Lee} 
\committee[2]{Won-Zoo Chung}
\committee[3]{Tae-Eui Kam}

\graduateDate{2026}{2}
\submitDate{2025}{10}
\approvalDate{2025}{12}

\captionLineSpacing{150}
\abstractLineSpacing{200}
\krAbstractLineSpacing{200}
\TOCLineSpacing{200}
\contentLineSpacing{200}
\acknowledgementLineSpacing{200}

\makeatletter
\let\makededicationpage\relax
\let\makepreface\relax

\makeatother

\begin{document}

\input{chapters/introduction}

\input{chapters/related_work}
\input{chapters/method}
\input{chapters/experiments}
\input{chapters/conclusion}

\end{document}

%% file: chapters/introduction.tex
\input{MAIN_Variables}

\chapter{Introduction}

\begin{figure}[htb!]
\centering
\includegraphics[width=1\linewidth]{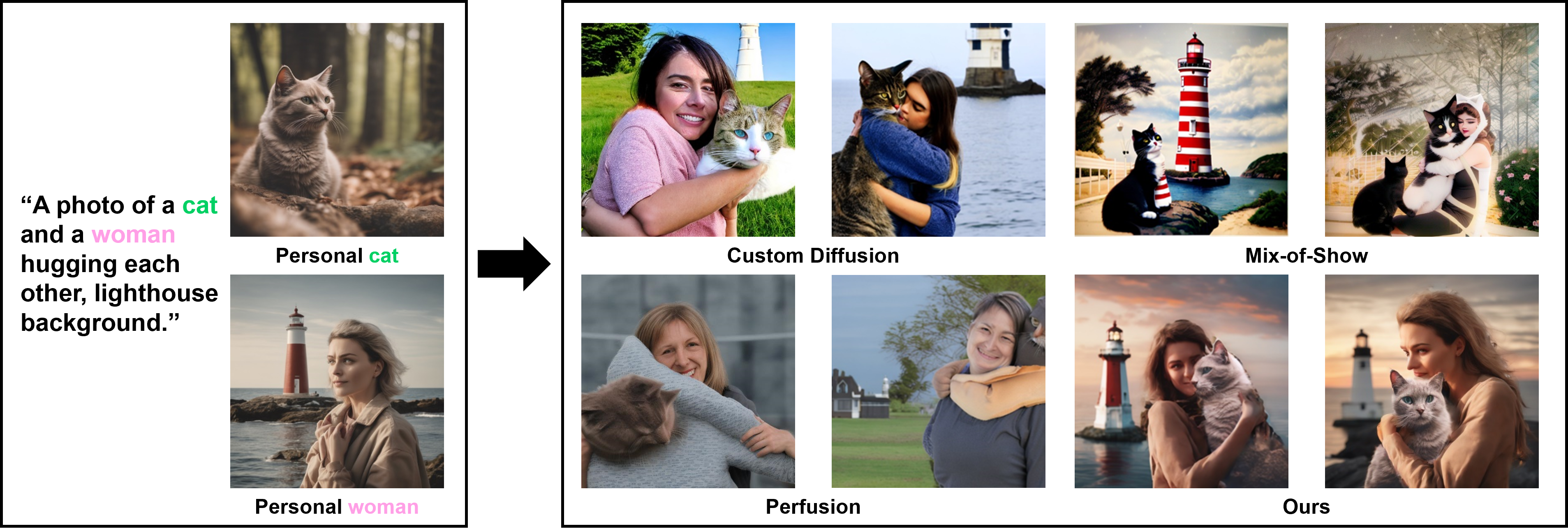}
\caption{
\textbf{Comparison of multi-subject personalization methods.}
Leftmost images show the personalized references for the cat and the woman. Remaining images depict the outputs of Custom Diffusion, Perfusion, Mix-of-Show, and our method (PnP-MIX), all conditioned on the prompt:
\textit{"A photo of a cat and a woman hugging each other, lighthouse background."}
While existing approaches often cause subjects to disappear, appearances to blend, or structural details to distort, our method preserves each concept’s features and the overall scene context, delivering coherent and high-quality results.
}
\label{fig:short}
\end{figure}

\sloppy
The field of Text-to-Image (T2I) generation has significantly advanced, enabling the creation of images aligned with personalized user-defined concepts. Nonetheless, effectively combining multiple distinct concepts within a single image remains a considerable challenge~\cite{min2022attentional}. This task requires careful coordination of diverse concepts, ensuring they align harmoniously~\cite{lim2000text} with the provided textual context while preserving their individual characteristics. Approaches such as Custom Diffusion~\cite{kumari2023multi} exemplify personalization methods that fine-tune cross-attention layers within pre-trained diffusion models by using multiple personalized reference images. Despite their strengths, tuning-based approaches~\cite{kumari2023multi, ruiz2023dreambooth, tewel2023key, gu2024mix} often encounter overfitting issues, significantly impacting their generalization ability, particularly when tasked with simultaneously integrating multiple personalized concepts into a coherent image. Consequently, recent research has shifted toward exploring tuning-free methodologies~\cite{alaluf2024cross, cao2023masactrl, nam2024dreammatcher}.

However, existing methods continue to struggle with effectively managing interactions among multiple objects, particularly in complex scenarios, as illustrated in Figure~\ref{fig:short}. They often cause unintended blending of personalized concepts and distortions or unwanted changes in non-personalized regions, undermining image integrity and quality.

To address persistent challenges in multi-concept image personalization, such as concept leakage, background distortion, and incorrect object interactions, we propose plug-and-play multi-concept adaptive blending for high-fidelity text-to-image synthesis (\textbf{PnP-MIX}). Our framework embeds multiple personalized concepts into a single image using only one reference image per concept, requires no additional tuning, and dramatically reduces both computational cost and the risk of overfitting.

PnP-MIX operates in two stages. In the first stage, we generate a background image and extract precise masks~\cite{ren2024grounded} for the concept insertion regions~\cite{li2023generate}. We then apply Edit-Friendly DDPM Inversion~\cite{huberman2024edit} to convert the background, its inpainted counterpart~\cite{yu2023inpaint}, and each personalized concept image into editing-ready latent representations.
In the second stage, these prepared latents and masks guide the image synthesis to seamlessly integrate multiple personalized concepts. To support this process, we introduce three key techniques.

First, we introduce \textbf{Guided appearance attention}, which dynamically adjusts the key and value vectors of each personalized concept based on its reference image. To better preserve each concept’s visual appearance during synthesis, our mechanism further reweights the value vectors, allowing the model to emphasize concept-specific visual cues more effectively. This strategy significantly enhances both the visual fidelity and prominence of each personalized concept in the generated image.
Second, we utilize \textbf{Mask-guided noise mixing}, which spatially combines noise predictions from the background and each personalized reference according to their respective masks. This selective fusion ensures that non-personalized (background) regions retain their original noise characteristics, while personalized object regions receive targeted noise injection only within their masks, thereby effectively preventing undesired blending during synthesis.
Finally, we propose \textbf{Background dilution++}, which dilates each personalized concept mask to create an expanded mask and, before self-attention, uses it to fuse the inpainted background latent (via Inpaint Anything~\cite{yu2023inpaint}) into non-personalized regions while preserving personalized latents. By weighting the background latent outside the dilated mask and maintaining the personalized features inside, this module minimizes concept leakage and improves boundary alignment between personalized concepts and their masks. The expanded mask is obtained by dilating the original object mask, ensuring that personalized attributes remain confined to their designated regions.

A preliminary version of this work appeared in~\cite{woo2025flipconcept}, where it demonstrated promising results but also presented notable limitations. Specifically, it revealed that, in self-attention–based concept insertion, misalignment occasionally occurred between the inserted concepts and the designated mask regions, leading to localized background distortions. Additionally, despite earlier background dilution~\cite{woo2025flipconcept}, concept leakage persisted among closely positioned personalized concepts. 
To address these issues, background dilution++ employs an inpainted background latent–based masking strategy that minimizes concept leakage and improves boundary alignment between personalized concepts and their masks.
Through extensive experimentation, including rigorous user studies and quantitative evaluations using CLIP~\cite{radford2021learning} and DINO~\cite{caron2021emerging,oquab2023dinov2} metrics, we demonstrate the robustness, visual fidelity, and contextual accuracy of images generated by PnP-MIX. Empirical results show that our method outperforms existing techniques, particularly in maintaining clear concept boundaries and background coherence under complex conditions.

In summary, the key contributions of this research include:
\begin{itemize}
  \item We identify key limitations in existing multi-concept personalization methods, namely concept leakage, background distortion, and incorrect object interactions.
  \item We propose PnP-MIX, a novel, tuning-free framework for integrating multiple personalized concepts within a single image, requiring no additional tuning and minimal computational overhead. By employing guided appearance attention, mask-guided noise mixing, and background dilution++, PnP-MIX effectively addresses the complexities inherent in multi-concept personalization.
  \item We validate the effectiveness of our approach through comprehensive empirical evaluations, including rigorous user studies and quantitative metrics (CLIP~\cite{radford2021learning} and DINO~\cite{caron2021emerging,oquab2023dinov2}), demonstrating superior visual and contextual quality compared to existing methods.
\end{itemize}

%% file: MAIN_Variables.tex

%% file: chapters/related_work.tex
\input{MAIN_Variables}
\chapter{Related Works \& Preliminaries}
\sloppy
\section{Related Works}

\subsection{Diffusion-based Text-to-Image Generation}
Denoising diffusion models~\cite{ho2020denoising,song2020denoising} and their advanced variants~\cite{rombach2022high,podell2023sdxl, saharia2022photorealistic, ramesh2022hierarchical} have significantly advanced Text-to-Image (T2I) generation. Utilizing robust text encoders such as CLIP~\cite{radford2021learning}, these models enable high-quality image synthesis guided by textual descriptions~\cite{lee1990translation}. Recent studies have introduced approaches for high-resolution images~\cite{rombach2022high}, classifier-free guidance for controllability~\cite{ho2022classifier}, and integrating larger-scale diffusion architectures~\cite{podell2023sdxl}.

\subsection{Image Editing with Diffusion Models}
Various editing techniques have been developed to leverage diffusion-based image generation.  
MasaCtrl~\cite{cao2023masactrl}, for example, demonstrates that freezing keys and values in self-attention layers within the denoising network~\cite{ronneberger2015u} effectively preserves an image’s original features. Subsequent methods adjusting self-attention layers to better maintain shapes~\cite{cao2023masactrl,nam2024dreammatcher,alaluf2024cross} allow for localized editing~\cite{hwang2006full} and global style transfer~\cite{zhang2023inversion} without extensive model retraining. Nevertheless, the efficacy of these approaches in scenarios involving multiple objects remains insufficiently validated~\cite{yang2007reconstruction}. While recent diffusion-based editing frameworks have achieved impressive results in various domain-specific contexts—including localized image editing~\cite{huang2025pfb}, facial reaction generation~\cite{li2025reactdiff}, and subject-driven video manipulation~\cite{zuo2025cut}—their generalization to arbitrary multi-concept or cross-domain scenarios remains underexplored.

\section{Preliminaries}

\subsection{Multi-Concept Personalization}
Generating images incorporating multiple personalized concepts remains challenging. Textual Inversion~\cite{gal2022image} optimizes text embeddings~\cite{voynov2023p+, li2023blip} to represent novel concepts, prompting further research involving full~\cite{ruiz2023dreambooth} or partial model fine-tuning~\cite{kumari2023multi, han2023svdiff}, and inserting rank-one edits~\cite{tewel2023key}. Concurrent training of multiple concepts often leads to concept blending or loss, spurring recent methods to mitigate these issues through weight merging~\cite{gu2024mix}. Training-free approaches such as Concept Weaver~\cite{kwon2024concept, kwon2024tweediemix} and DreamMatcher~\cite{nam2024dreammatcher} allow concept combination during sampling but may still require inversion or fine-tuning at inference. Our approach dramatically reduces additional optimization and flexibly generates superior multi-concept images, effectively extending prior work.

\subsection{Latent Diffusion Models}
Our method leverages the Latent Diffusion Model (LDM)~\cite{rombach2022high}, operating in a compressed latent space instead of directly in high-dimensional image spaces, thereby achieving computational efficiency and high-resolution outputs. An autoencoder encodes images into latent vectors, which a decoder then reconstructs. Text prompts are encoded into semantic embeddings using a pretrained CLIP text encoder~\cite{radford2021learning}. These text embeddings serve as conditions \(c\) to ensure that generated images match the input descriptions, while images themselves are also mapped to the compressed latent space via an autoencoder. The LDM training minimizes:

\begin{equation}
L = \mathbb{E}_{x_0, \epsilon \sim \mathcal{N}(0,I)}\left[\|\epsilon - \epsilon_\theta(x_t, t, c)\|_2^2\right],
\end{equation}

\noindent where \(x_t\) denotes the noise-perturbed image at time step \(t\), and \(\epsilon_\theta\) denotes the noise estimation function parameterized by \(\theta\)~\cite{ho2020denoising}. This forms the basis for text-conditioned image generation.

\subsection{Self-Attention in Stable Diffusion}
We employ Stable Diffusion (SD)~\cite{rombach2022high}, an LDM-based model featuring a denoising U-Net architecture~\cite{ronneberger2015u,ding2024freecustom}. SD includes residual blocks and self-attention and cross-attention modules distributed across seven blocks and sixteen layers, enabling efficient latent-space image generation. The self-attention mechanism particularly influences the layout and detailed content of generated images. At each time step \(t\), noisy latent codes \(z_t\) produce query (Q), key (K), and value (V) vectors from intermediate representations \(\phi_l(z_t)\), via linear projections \(W_Q, W_K, W_V\):

\begin{equation}
Q = W_Q(\phi_l(z_t)), \quad K = W_K(\phi_l(z_t)), \quad V = W_V(\phi_l(z_t)).
\end{equation}

Attention scores, calculated as the inner products between queries and keys divided by \(\sqrt{d_k}\), are normalized using a softmax function to maintain numerical stability and improve training efficiency. Updated features \(\Delta \phi(i,j)\) are obtained by summing weighted value vectors:

\begin{equation}
A(i,j) = \mathrm{softmax}\left(\frac{q_{(i,j)} \cdot K^T}{\sqrt{d_k}}\right), \quad \Delta \phi(i,j) = A(i,j) \cdot V.
\end{equation}

Thus, the model effectively highlights regions in images relevant to the textual semantics. Prior analyses~\cite{alaluf2024cross,cao2023masactrl,nam2024dreammatcher} demonstrate that self-attention in diffusion models tends to capture both global structure (e.g., object arrangement and shape) and local appearance attributes~\cite{roh2007accurate} (e.g., color and texture), and that these aspects can be controlled or emphasized independently depending on downstream modifications.

%% file: chapters/method.tex
\chapter{Methods}
\sloppy
\begin{figure}[htb!] 
\centering
\includegraphics[width=1\linewidth]{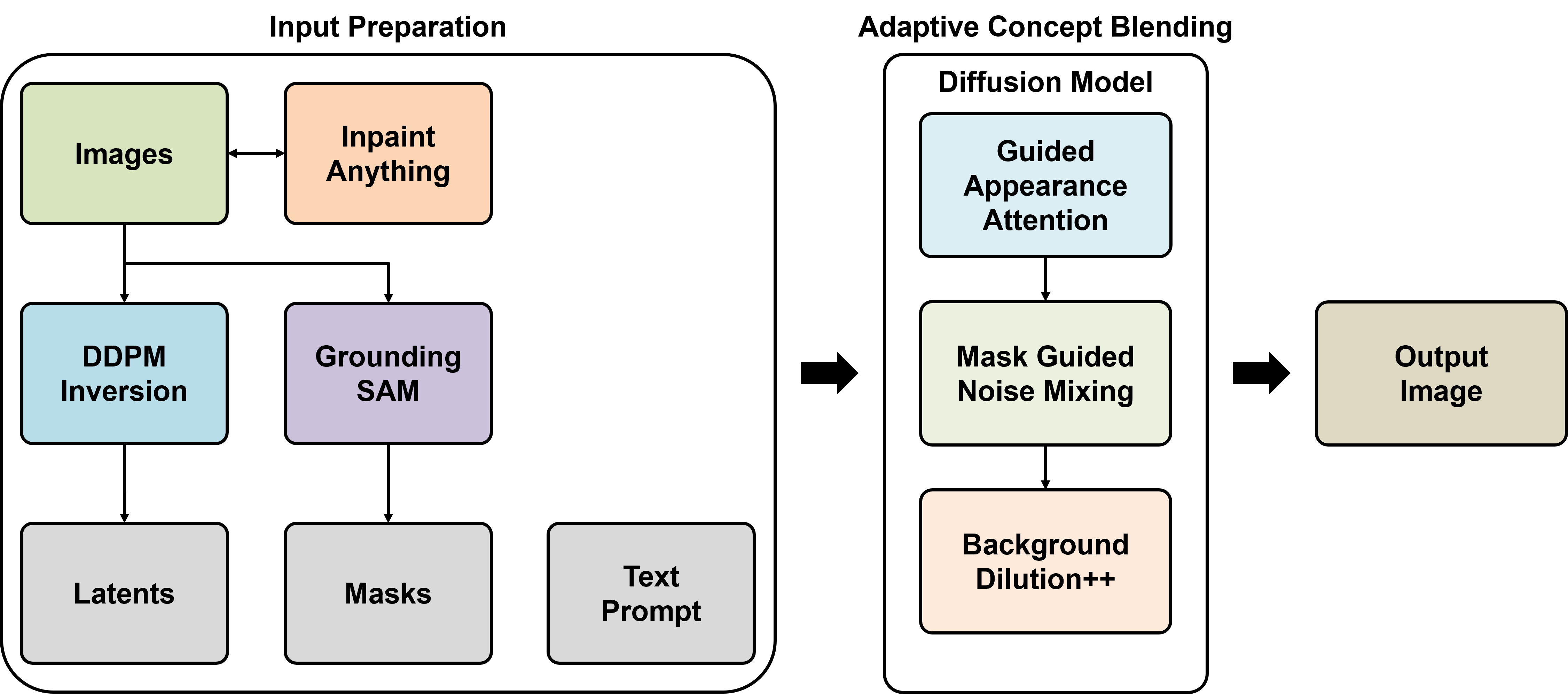}
\caption{
\textbf{Overview of PnP-MIX.}
Our framework integrates multiple personalized concepts into a single image via a two-stage pipeline.
In the first stage, background and object masks are extracted using Grounded SAM~\cite{ren2024grounded}, and Inpaint Anything~\cite{yu2023inpaint} is used to generate a background image with the original object regions removed.
In the second stage, latent representations of the inpainted background, background, and each personalized concept, along with their masks, are obtained via DDPM Inversion~\cite{huberman2024edit} and fused using guided appearance attention, mask-guided noise mixing, and Background Dilution++.
Unlike tuning-based personalization methods that rely on random background generation, PnP-MIX supports explicit background selection, providing fine-grained control over scene composition.
}
\label{fig}
\end{figure}

In this section, we introduce \textbf{PnP-MIX}, a framework that integrates user-specified concepts into a single image using a pretrained diffusion model with no extra fine-tuning.
It seamlessly merges diverse concepts into a coherent image, producing high-quality results that closely align with the user’s intent.
This approach overcomes the limitations of traditional object-editing methods by enabling richer and more complex scenarios, thereby extending the model’s applicability to simultaneous handling of multiple concepts.

Motivated by the sequential image generation approach of Kwon et al.~\cite{kwon2024concept}, PnP-MIX consists of two key stages.
In the first stage, we select a background image that matches the prompt and apply a text-driven segmentation model~\cite{ren2024grounded} to extract both object and background masks~\cite{ahmad2006human}. We then employ an inpainting model~\cite{yu2023inpaint} to generate a background image with the personal concept regions removed. Subsequently, the concept-removed background image, the original background image, and each personal concept image are encoded into latent space via DDPM Inversion~\cite{huberman2024edit}, and the background latent representation is duplicated for later fusion.
In the second stage, all latents and masks are jointly processed by the diffusion model to produce the final output image \(I_{\text{out}}\), seamlessly integrating the user’s personal concepts into the chosen scene. Figure~\ref{fig} illustrates the overall pipeline, and each component is described in detail in the following sections.

\section{Input Preparation}

\subsection{Background Image and Mask Generation}

\begin{figure}[htbp]
\centering
\includegraphics[width=1\linewidth]{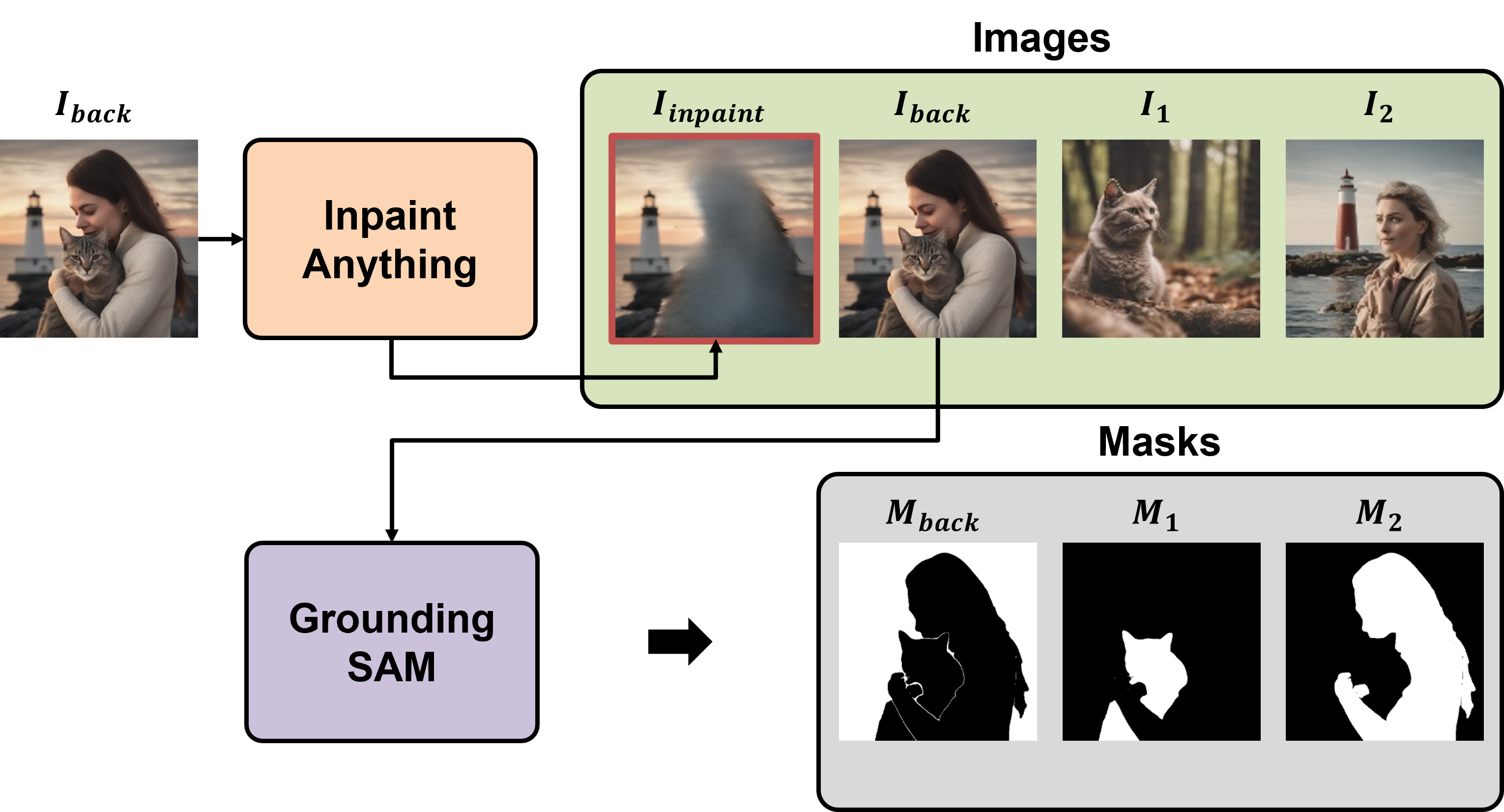}
\caption{
\textbf{Input preparation pipeline.}
Given a user-selected background image \(I_{\mathrm{back}}\) containing the target concept, we apply Inpaint Anything~\cite{yu2023inpaint} to remove the specified object region and obtain the inpainted background \(I_{\mathrm{inpaint}}\). Then, Grounded SAM~\cite{ren2024grounded} generates binary masks \(M_{\mathrm{back}}\), \(M_{1}\), and \(M_{2}\) to segment the background and each personalized concept. These components constitute the inputs fed into the PnP-MIX framework.}
\label{fig:input}
\end{figure}

Most prior methods~\cite{kumari2023multi, gal2022image, tewel2023key, gu2024mix} generate images by randomly filling all regions outside the personal concept areas. 
In contrast, we first sample multiple candidate backgrounds using Stable Diffusion XL~\cite{podell2023sdxl} and allow users to directly select a scene that both explicitly contains the personal concepts and aligns with their preferences.

Once a background \(I_{\mathrm{back}}\) is selected, we extract masks \(M_{1}, M_{2}, \dots\) for each personal concept and a background mask \(M_{\mathrm{back}}\) using Grounded SAM~\cite{ren2024grounded,liu2024grounding,kirillov2023segment}.
We then apply Inpaint Anything~\cite{yu2023inpaint} to remove the personal concept regions from \(I_{\mathrm{back}}\), producing an inpainted background \(I_{\mathrm{inpaint}}\). The resulting images \(\{I_{\mathrm{back}}, I_{\mathrm{inpaint}}\}\) and masks \(\{M_{\mathrm{back}}, M_{1}, M_{2}\}\) serve as inputs to the subsequent fusion process.

\subsection{Image Inversion}

\begin{figure}[htbp]
\centering
\includegraphics[width=1\linewidth]{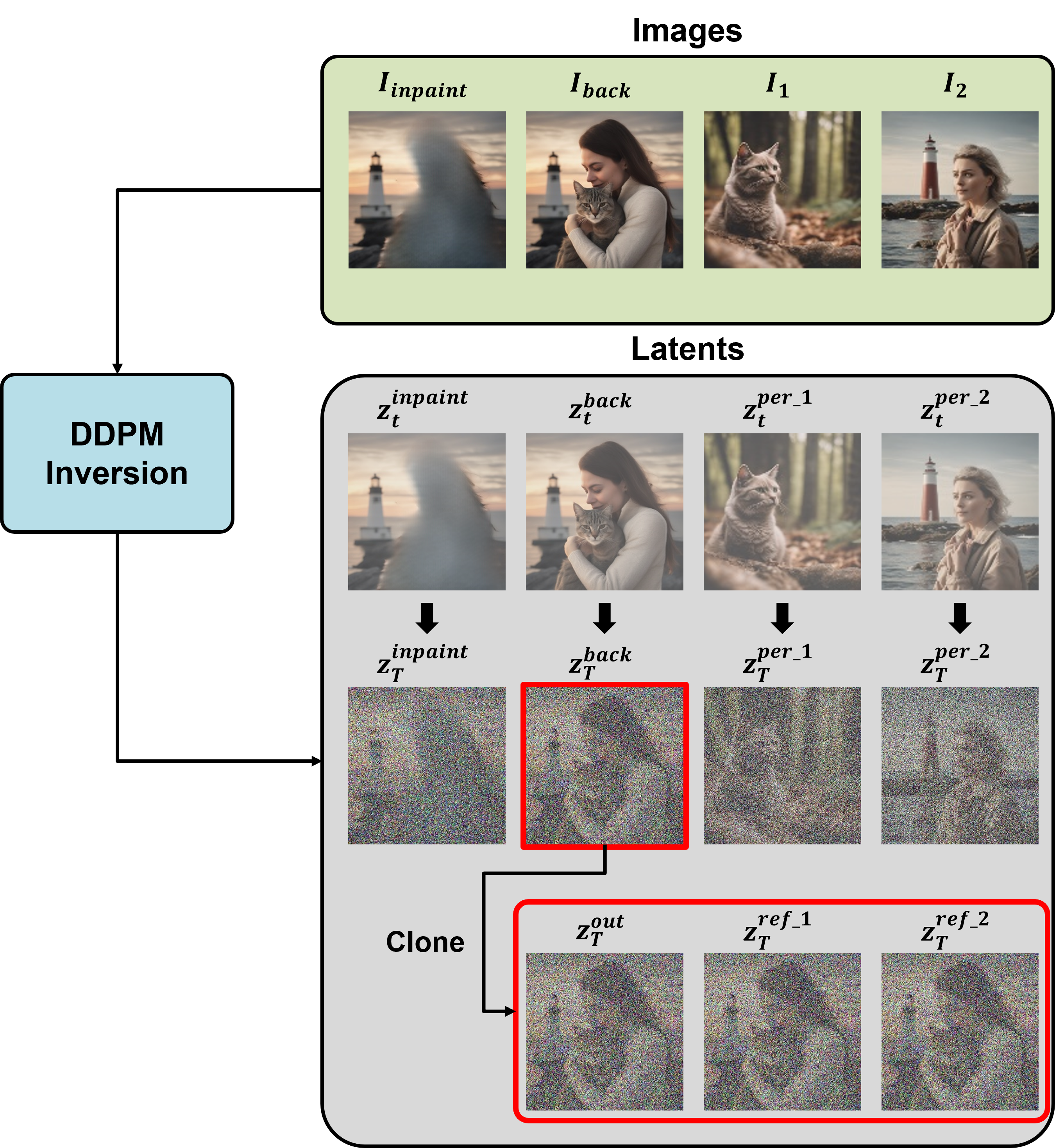}
\caption{
\textbf{Edit-Friendly DDPM Inversion and latent cloning.}
We apply Edit-Friendly DDPM Inversion~\cite{huberman2024edit} to the inpainted background \(I_{\mathrm{inpaint}}\), the selected background \(I_{\mathrm{back}}\), and the personalized concept images \(I_{1}\), \(I_{2}\), to obtain their noise codes \(z_{T}^{\mathrm{inpaint}}\), \(z_{T}^{\mathrm{back}}\), \(z_{T}^{\mathrm{per}_1}\), and \(z_{T}^{\mathrm{per}_2}\). The background latent \(z_{T}^{\mathrm{back}}\) (highlighted in red) is then cloned to produce the output latent \(z_{T}^{\mathrm{out}}\) and reference latents \(z_{T}^{\mathrm{ref}_1}\), \(z_{T}^{\mathrm{ref}_2}\). These duplicated latents guide the subsequent multi-concept fusion stage, ensuring faithful background reconstruction and accurate embedding of each personalized concept.
}
\label{fig:ddpm}
\end{figure}

After the background, inpainted background, and personal concept images are selected, we extract their latent representations.  
Traditional methods employ DDIM Inversion~\cite{song2020denoising, han2023improving}, which reverses the diffusion process to recover latent codes. However, such methods require numerous steps, leading to reduced reconstruction accuracy and loss of fine-grained details. To overcome these limitations, we apply Edit-Friendly DDPM Inversion~\cite{huberman2024edit} to extract noise latents that directly reflect the original images, thereby preserving structural integrity and detail while supporting diverse editing tasks.

Edit-Friendly DDPM Inversion constructs a sequence based on a probabilistic path originating from the original image \(x_0\) (e.g., \(I_{\mathrm{inpaint}}, I_{\mathrm{back}}, I_{1}, I_{2}\) in Figure~\ref{fig:input}). Specifically, at each timestep \(t\) (\(1 \le t \le T\)), noise \(\tilde{\epsilon}_t \sim \mathcal{N}(0, I)\) is independently sampled, and \(x_t\) is computed as follows:
\begin{equation}
x_t = \sqrt{\bar{\alpha}_t}\, x_0 + \sqrt{1 - \bar{\alpha}_t}\, \tilde{\epsilon}_t,
\end{equation}
\noindent where \(\bar{\alpha}_t\) is the cumulative noise schedule parameter. Independent sampling of noise \(\tilde{\epsilon}_t\) at each timestep increases the divergence between successive latent representations \(x_t\) and \(x_{t-1}\), thereby enhancing the variance and embedding fine-grained details more accurately in the resulting noise latents \(z_T, \dots, z_1\). This stronger variance embeds the original image’s structure more faithfully than conventional DDPM sampling~\cite{ho2020denoising}.
The auxiliary sequence \((x_1, \dots, x_T)\) produced by Edit-Friendly DDPM Inversion is then used to compute the noise map. At each timestep \(t\), given \(\hat{\mu}_t(x_t)\) and \(\sigma_t\), the noise latent \(z_t\) is obtained as:
\begin{equation}
z_t = \frac{x_{t-1} - \hat{\mu}_t(x_t)}{\sigma_t}.
\end{equation}

This inversion retraces the model’s forward generation process, yielding a noise map that closely reflects the original image’s features. As a result, it preserves fine-grained details while providing a versatile latent code for editing. These latents play a crucial role in later editing stages, enabling seamless integration of new objects or concepts without sacrificing structural integrity.

Among the extracted latents, the background latent \(z_T^{\mathrm{back}}\) is duplicated so that the number of background clones exceeds the number of personal concepts by one (e.g., one output latent \(z_T^{\mathrm{out}}\) and reference latents \(z_T^{\mathrm{ref}_1}\), \(z_T^{\mathrm{ref}_2}\) as shown in Figure~\ref{fig:ddpm}). The clone labeled \(z_T^{\mathrm{out}}\) is used to generate the final output image \(I_{\mathrm{out}}\), while the remaining clones \(z_T^{\mathrm{ref}_1}, z_T^{\mathrm{ref}_2}, \dots\) serve as reference latents. These reference latents ensure reliable embedding of each targeted concept during fusion. 
In parallel, the latent \(z_T^{\mathrm{inpaint}}\) obtained from the inpainted background \(I_{\mathrm{inpaint}}\) acts as a visual prior for regions where the original object was removed. By incorporating this inpainted latent during background 
dilution++, PnP-MIX further suppresses unintended concept leakage into background areas, reinforcing spatial consistency and semantic cleanliness.

Finally, all prepared latents and masks are passed into the Stable Diffusion model~\cite{rombach2022high} to generate the final image \(I_{\mathrm{out}}\), which integrates multiple personal concepts into a unified scene.

\section{Adaptive Concept Blending}

Once all inputs are ready, we feed them into the diffusion model, incorporating three core techniques \textbf{Guided appearance attention}, \textbf{Mask-guided noise mixing}, and \textbf{Background dilution++} to handle multi-concept blending challenges. These approaches maintain the scene’s structural coherence while accurately conveying appearance details, facilitating smooth integration of concepts and producing high-quality images.

\begin{figure}[htbp]
\centering
\includegraphics[width=1\linewidth]{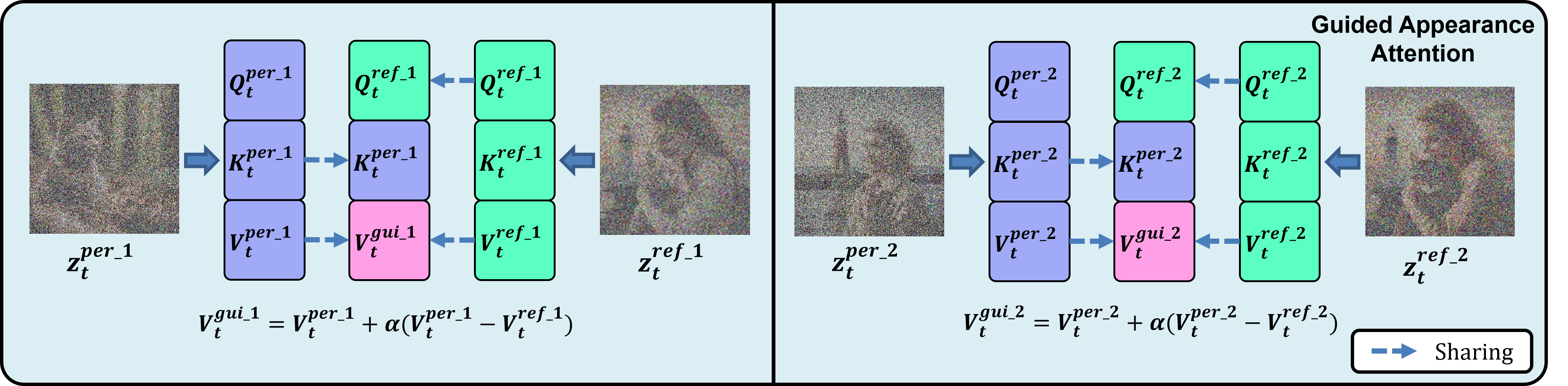}
\caption{\textbf{Guided appearance attention.} At each self-attention layer, the personal concept latent $z_t^{\mathrm{per}}$ and the corresponding reference latent $z_t^{\mathrm{ref}}$ are processed to inject appearance information from the personal concept into the reference, while preserving spatial structure. The key from $z_t^{\mathrm{ref}}$ is replaced with that from $z_t^{\mathrm{per}}$, while the value is adjusted using value guidance. This mechanism enriches the reference latent with vivid color and texture cues from the personal concept while preserving its spatial structure.}
\label{fig:gaa}
\end{figure}

\textbf{Guided appearance attention} injects appearance details by replacing the key of the reference image with the key from the personal concept image and guiding the reference value inspired by classifier-free guidance~\cite{ho2022classifier}, which highlights the distinctive appearance characteristics of the personalized concept. In this context, recent studies~\cite{alaluf2024cross, cao2023masactrl, nam2024dreammatcher} have analyzed the self-attention layer architecture in the denoising U-Net of text-to-image diffusion models and proposed that fixing the key and value preserves an object’s visual features even under non-structural transformations. In particular, self-attention can be split into a structure path and an appearance path, where the appearance path is closely tied to visual properties such as color and texture and plays a pivotal role in enhancing image quality. Based on this idea, guided appearance attention effectively transfers the appearance information of the personal concept image while preserving the structural information of the reference image within the self-attention layer. This enhances text–image alignment while improving overall visual quality.
The value guidance is defined as:
\begin{equation}
V_t^{\text{gui}\_{\mathit{i}}} = V_t^{\text{per}\_{\mathit{i}}} + \alpha \cdot \bigl(V_t^{\text{per}\_{\mathit{i}}} - V_t^{\text{ref}\_{\mathit{i}}}\bigr),
\end{equation}
\noindent  where \(V_t^{\text{per}_i}\) and \(V_t^{\text{ref}_i}\) denote the value vectors corresponding to the \(i\)-th personal concept latent \(z_t^{\text{per}_i}\) and the reference latent \(z_t^{\text{ref}_i}\), respectively, and \(\alpha\) is a guidance scale controlling the strength of appearance injection. In our experiments, we set \(\alpha=0.15\) to balance visual quality and alignment, but this value can be adjusted to emphasize appearance attributes as needed. The reference latent’s key and value after merging with the personal concept’s key and value are then used in the standard self-attention operation~\cite{cao2023masactrl, alaluf2024cross}, injecting personal concept appearance into the reference’s structural map. Finally, the refined latents are forwarded to the diffusion model’s denoising stage.

\begin{figure}[htb!]
\centering
\includegraphics[width=1\linewidth]{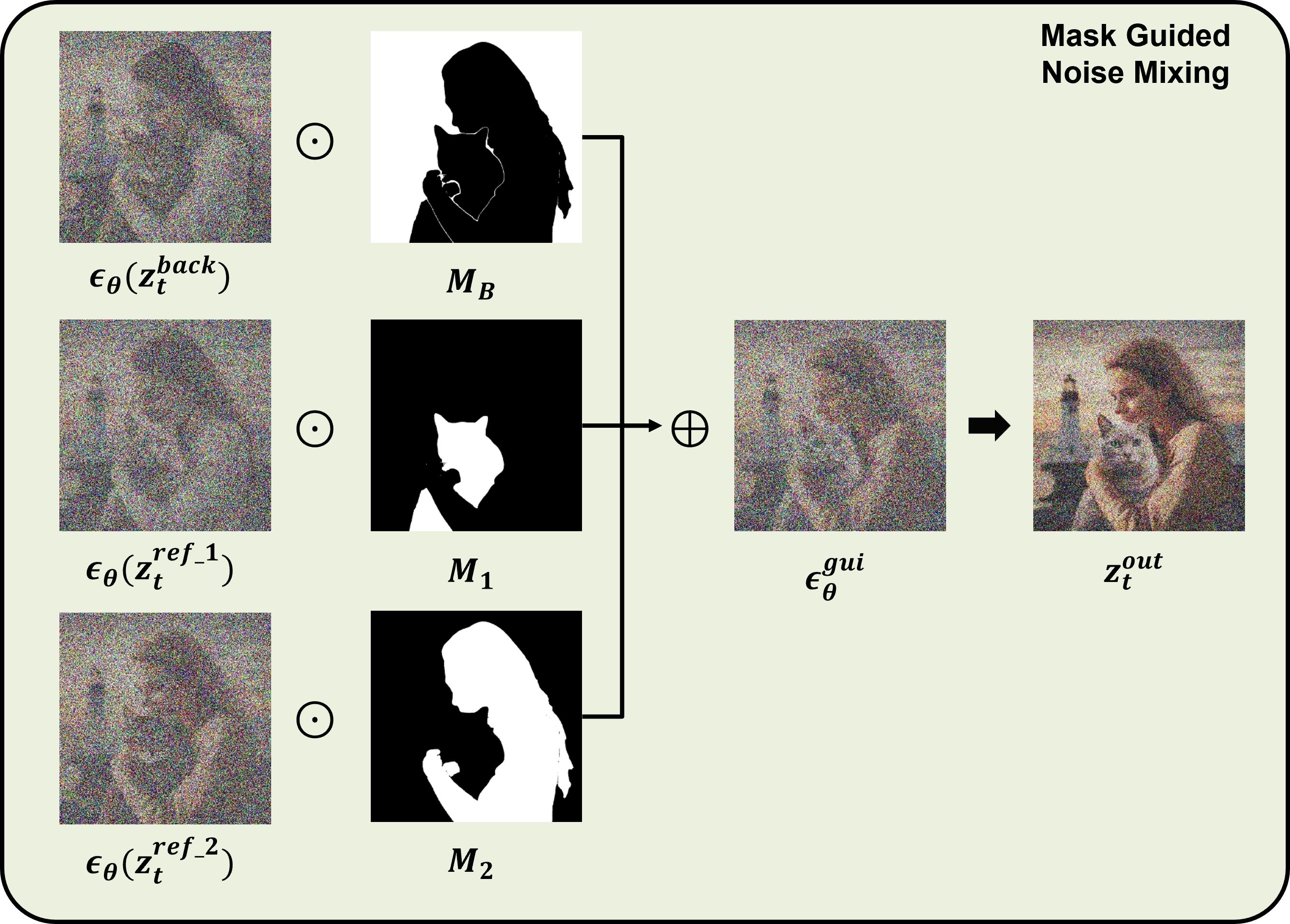}
\caption{
\textbf{Mask-guided noise mixing.} 
At each timestep \(t\), the diffusion model predicts noise maps for the background latent \(\epsilon_\theta(z_t^{\mathrm{back}})\) and for each reference latent \(\epsilon_\theta(z_t^{\mathrm{ref}_i})\). Each noise map is then multiplied element-wise by its corresponding mask \(M_B\) for the background and \(M_i\) for the \(i\)-th object region. The masked noise maps are summed to form the guided noise \(\epsilon_\theta^{\mathrm{gui}}\), ensuring that each personal concept is applied strictly within its designated region while faithfully preserving background details.
}
\label{fig:mgnm}
\end{figure}

\textbf{Mask-guided noise mixing} directly applies masks at the noise operation stage, overcoming limitations of previous approaches and enabling a stable multi-concept blend by separating objects from the background. Previous studies~\cite{kumari2023multi, tewel2023key, gu2024mix, hertz2022prompt} attempted to integrate concepts using regional sampling or cross-attention mechanisms within the bottleneck layer, but the low resolution at that stage made fine-grained manipulation difficult and often led to unintended changes. 
Mask-guided noise mixing blends noise at the mask level to ensure that each personalized concept is applied strictly within its designated region. Specifically, the object mask \(M_i\) identifies the editing region, while the background mask \(M_B\) covers the area to be preserved. Noise integration is performed via a Hadamard product ($\odot$) as follows:
\begin{equation}
\epsilon_\theta^{\text{gui}} = \epsilon_\theta(z_t^{\text{back}}) \odot M_B + \sum_{i=1}^n \epsilon_{\theta}(z_t^{\text{ref}\_{\mathit{i}}}) \odot M_i,
\end{equation}
\noindent where \(\epsilon_\theta(z_t^{\text{back}})\) denotes the noise predicted for the background latent and \(M_B\) is its associated mask. Likewise, \(\epsilon_{\theta}(z_t^{\text{ref}_i})\) and \(M_i\) correspond to the noise and mask for the \(i\)-th reference concept. The resulting guided noise \(\epsilon_\theta^{\text{gui}}\) is used in the denoising step~\cite{ho2020denoising,huberman2024edit} to generate the final output latent \(z_t^{\text{out}}\) at time step \(t\).
In this step, we also update each \(\,i\)-th reference latent by re-synthesizing non-object regions with background noise, using the corresponding object mask to identify which areas to substitute. Moreover, a swap guidance mechanism~\cite{alaluf2024cross} is employed to steer the denoising process toward the desired appearance, reducing distortions and enhancing realism. As a result, object and background regions remain distinctly separated, yet various personalized concepts from the text prompt are smoothly fused into a coherent image.

\begin{figure}[htb!]
\centering
\includegraphics[width=1\linewidth]{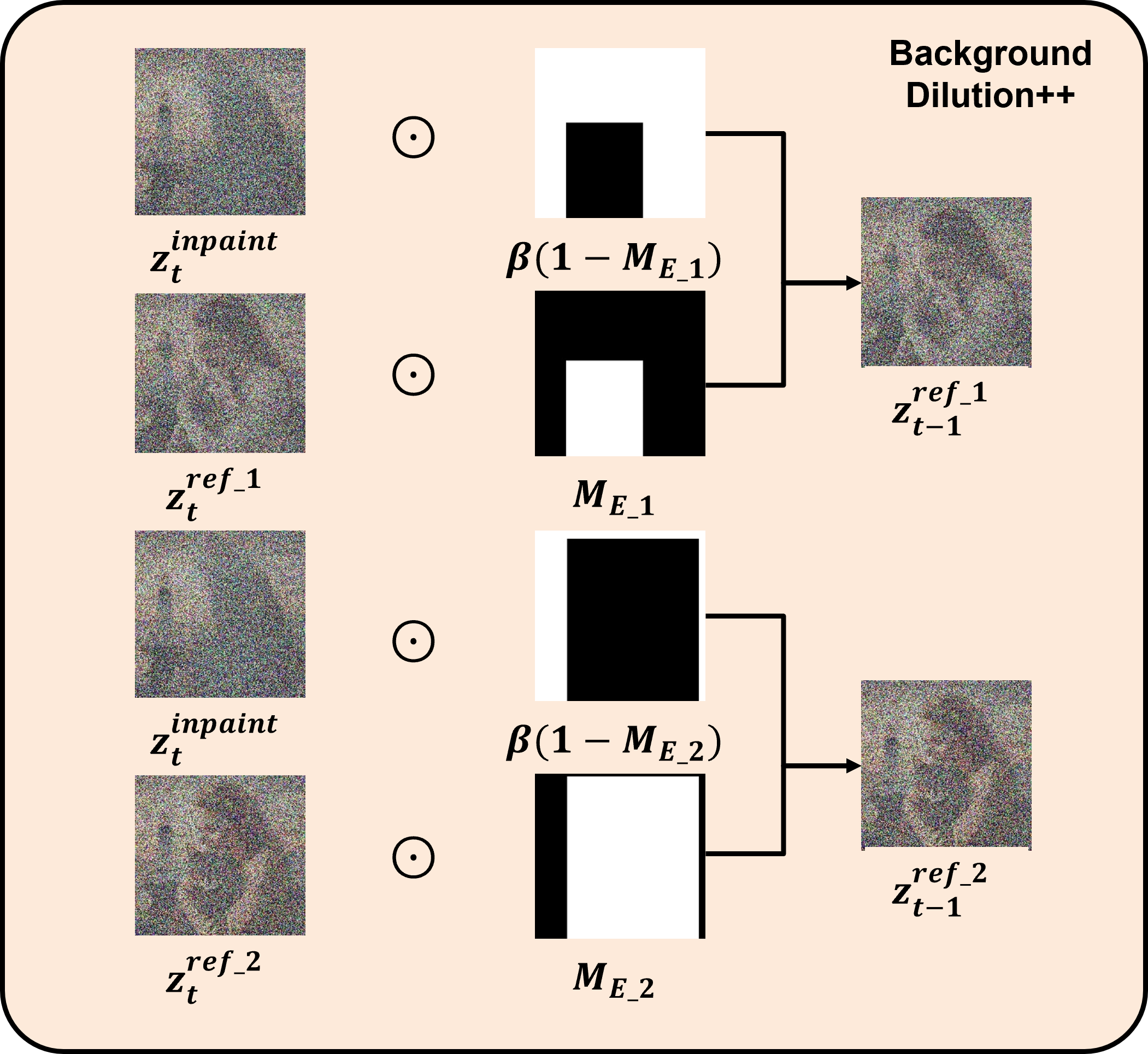}
\caption{\textbf{Background dilution++.} For each personal concept, we generate an expanded rectangular mask \(M_{E_i}\) around the object region. The inpainted background latent \(z_t^{\mathrm{inpaint}}\) is gated by \(\beta(1 - M_{E_i})\) and the reference latent \(z_t^{\mathrm{ref}_i}\) by \(M_{E_i}\), then summed to yield the updated latent \(z_{t-1}^{\mathrm{ref}_i}\).}
\label{fig:bd}
\end{figure}

\textbf{Background dilution++} reduces the influence of background within each reference latent during diffusion by utilizing object masks. Specifically, after generating the reference latent (but before subsequent steps), we create a minimal bounding rectangular mask \(M_E\) based on the extracted object mask. This mask is used to mix the inpainted background latent \(z_t^{\text{inpaint}}\) and the reference latent \(z_t^{\text{ref}}\), with the background latent weighted by a factor \(\beta\) to control its contribution and effectively dilute the background. The latent update equation is:
\begin{equation}
z_{t-1}^{\text{ref}\_{\mathit{i}}} = z_t^{\text{inpaint}} \odot \beta \bigl(1 - M_E\bigr) + z_t^{\text{ref}\_{\mathit{i}}} \odot M_E,
\end{equation}
\noindent where \(z_t^{\text{inpaint}}\) denotes the inpainted background latent~\cite{yu2023inpaint}, \(z_t^{\text{ref}_i}\) represents the reference latent for the \(i\)-th concept, and \(\beta\) (the dilution scale) adjusts their relative contributions. Through experimentation, we found \(\beta=0.8\) to be optimal. Additionally, \(M_E\) is an expanded bounding rectangular mask derived from the original object mask, which accurately isolates the region surrounding the target object.
This approach mitigates previously reported image distortion and concept leakage in multi-object scenarios. By reducing non-target background influence, background dilution++ prevents unwanted attribute propagation during self-attention, thereby preserving each concept’s intended representation more accurately.

By integrating these techniques, PnP-MIX efficiently synthesizes multiple personalized concepts into a single image, significantly improving image quality and textual fidelity while addressing common issues such as distortion and concept leakage.

%% file: chapters/experiments.tex
\chapter{Experiments}
\sloppy
\begin{figure}[htb!]
\centering
\includegraphics[width=1\textwidth]{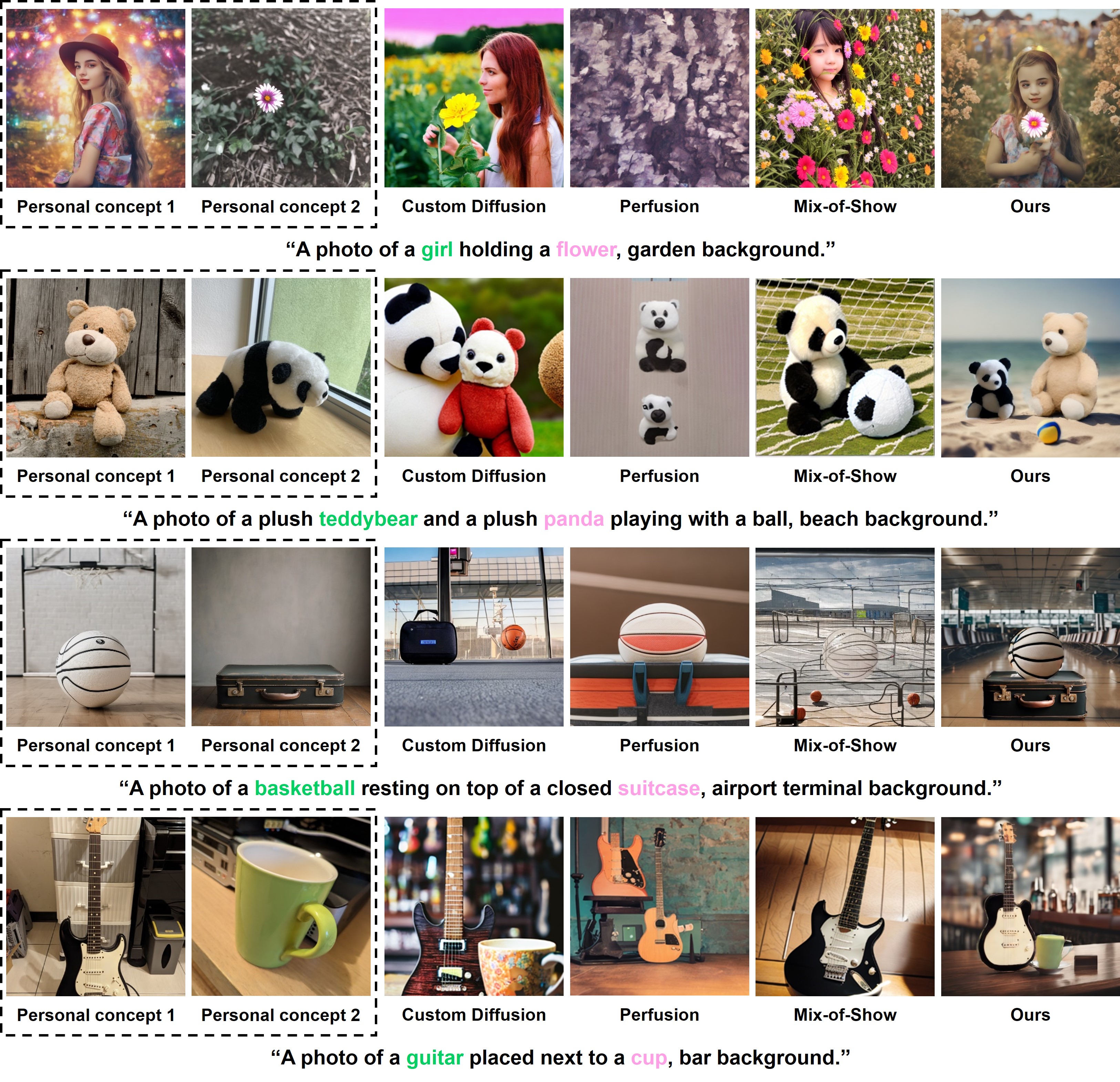}
\caption{\textbf{Qualitative evaluation of multi-concept personalization.} We compare our method PnP-MIX against baseline approaches (Custom Diffusion~\cite{kumari2023multi}, Perfusion~\cite{tewel2023key}, Mix-of-Show~\cite{gu2024mix}) using prompts that incorporate different pairs of personalized concepts (e.g., girl and flower) across various backgrounds.
Our method generates coherent results in these complex scenarios without undesired blending or loss of structural integrity.}
\label{fig3}
\end{figure}

\section{Experimental Settings}

All experiments were conducted using the Stable Diffusion v2.1 base model~\cite{rombach2022high}, generating 512×512-pixel images on two NVIDIA A40 GPUs. We adopted evaluation protocols from previous multi-personalized-concept studies~\cite{kumari2023multi, tewel2023key, kwon2024concept}, employing the Custom Concept 101 dataset~\cite{kumari2023multi} and additional samples synthesized with Stable Diffusion~\cite{rombach2022high}. Additionally, we constructed ten distinct evaluation sets using ChatGPT~\cite{achiam2023gpt} to cover various object interactions.

For quantitative analysis, we used two CLIP-based metrics, $T_\text{CLIP}$~\cite{radford2021learning} and $I_\text{CLIP}$~\cite{radford2021learning}, and one DINOv2-based metric, $I_\text{DINO}$~\cite{caron2021emerging,oquab2023dinov2}, all computed as cosine similarities. Specifically, $T_\text{CLIP}$ measures prompt–image consistency, while $I_\text{CLIP}$ and $I_\text{DINO}$ evaluate individual concept fidelity via regional feature embeddings. All evaluation conditions were kept uniform across methods to ensure fairness and reproducibility.
We benchmarked PnP-MIX against three baselines: Custom Diffusion~\cite{kumari2023multi}, Perfusion~\cite{tewel2023key}, and Mix-of-Show~\cite{gu2024mix}. PnP-MIX consistently outperformed these approaches in quantitative metrics, qualitative assessments, user preference evaluations, and practical runtime efficiency. Furthermore, we performed ablation studies to quantify each component’s contribution within our framework.

\begin{table}[htbp]
\caption{\textbf{Quantitative comparison of multi-personalized concept generation.} Our method (d) achieves the highest scores in $T_\text{CLIP}$, $I_\text{CLIP}$, and $I_\text{DINO}$~\cite{caron2021emerging,oquab2023dinov2}, reflecting its stronger correspondence with the prompt and more accurate concept representation.}
\centering
\resizebox{\linewidth}{!}{
\begin{tabular}{clccc}
\toprule
& \textbf{Method} & $T_\text{CLIP} \uparrow$ & $I_\text{CLIP} \uparrow$ & $I_\text{DINO} \uparrow$ \\
\midrule
(a) & Custom Diffusion~\cite{kumari2023multi}          & 0.3480 & 0.8078 & 0.4657  \\
(b) & Perfusion~\cite{tewel2023key}                & 0.3044 & 0.7616 & 0.3267  \\
(c) & Mix-of-Show~\cite{gu2024mix}      & 0.3206 & 0.7920 & 0.4262 \\
(d) & \textbf{PnP-MIX (Ours)}                           & \textbf{0.3823} & \textbf{0.8841} & \textbf{0.6542}  \\
\bottomrule
\end{tabular}
}
\label{tab:component_comparison}
\end{table}

\section{Generation Results}
\textbf{Qualitative and quantitative evaluation.} To evaluate the effectiveness of our proposed method, we generated 50 images per approach for both qualitative and quantitative assessments. Our results consistently highlight our method’s capability to produce coherent, visually appealing, and structurally precise multi-concept images.

Table~\ref{tab:component_comparison} summarizes the quantitative performance, measured by CLIP-based~\cite{radford2021learning} and DINO-based~\cite{caron2021emerging,oquab2023dinov2} metrics. Our approach achieves the highest scores across all three metrics: $T_\text{CLIP}$, $I_\text{CLIP}$, and $I_\text{DINO}$. These results indicate strong alignment between generated images and input prompts, as well as accurate representation of individual concepts. In contrast, baseline methods demonstrate performance deterioration as the number of concepts increases. For example, Custom Diffusion, despite explicitly supporting multiple concepts, frequently encounters structural inconsistencies and concept blending issues~\cite{lee2020uncertainty}, thus yielding lower quantitative scores compared to ours.

In addition to these quantitative results, Figure~\ref{fig3} provides qualitative comparisons using various challenging prompts involving multiple personalized concepts. Our method effectively manages intricate interactions between concepts, such as the example prompt \textit{"A photo of a basketball resting on top of a closed suitcase, airport terminal background."} Baseline methods commonly struggle to preserve distinct visual details and structural integrity. Specifically, Perfusion often has difficulty maintaining global coherence, and Mix-of-Show also frequently blends multiple concepts into ambiguous shapes. Conversely, our method clearly integrates each personalized concept, ensuring visual clarity and prompt fidelity.

\begin{figure}[htb!]
\centering
\includegraphics[width=1\textwidth]{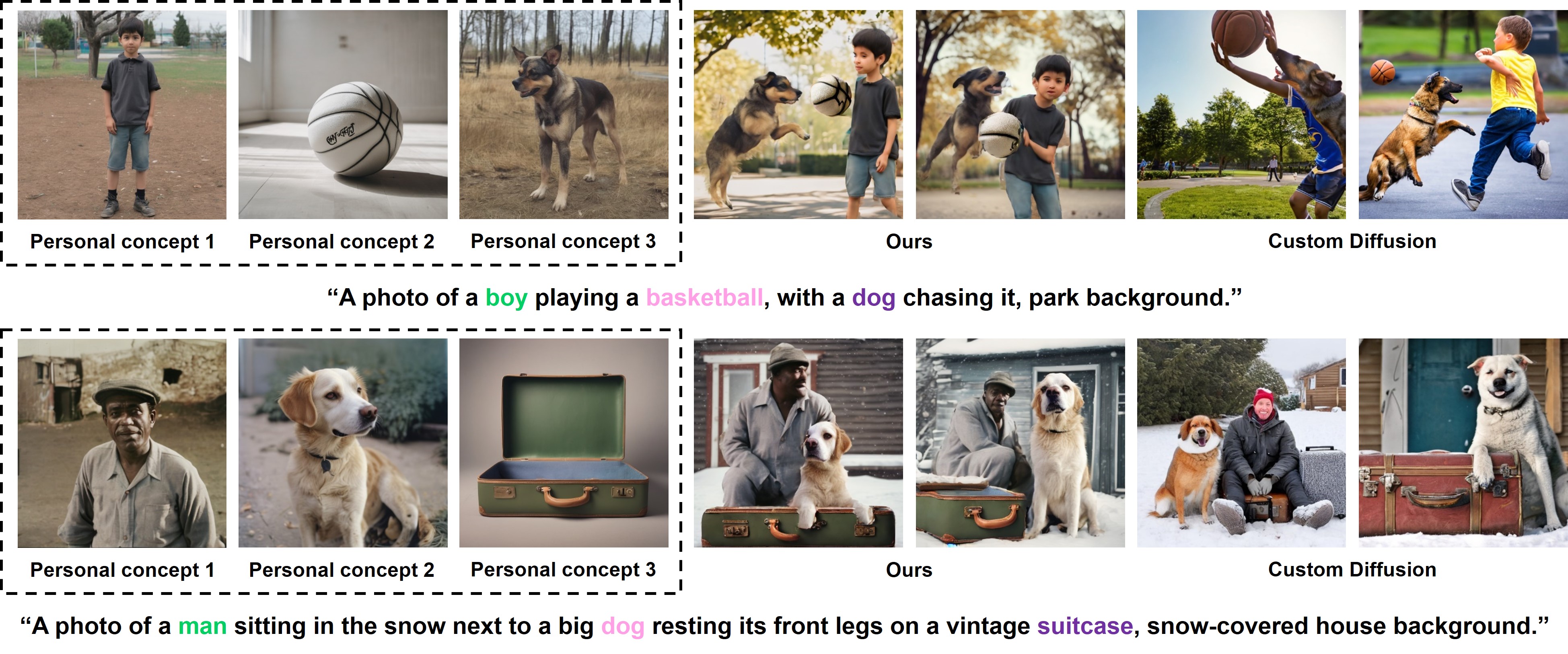}
\caption{\textbf{Robust multi-concept personalization in challenging scenarios.} We compare our approach against Custom Diffusion~\cite{kumari2023multi} using prompts that involve three interacting concepts. Custom Diffusion often misses or mixes elements, breaking the intended scene layout. In contrast, PnP-MIX reliably maintains all personalized concepts and preserves their spatial relationships.}
\label{fig:3concept}
\end{figure}

Moreover, Figure~\ref{fig:3concept} demonstrates that our method consistently outperforms Custom Diffusion~\cite{kumari2023multi} in scenarios involving three interacting concepts. Although Custom Diffusion incorporates multiple concepts through joint or separate fine-tuning, image quality notably deteriorates with increasing complexity due to representation and balancing challenges. In contrast, our method maintains spatial integrity and distinctiveness of each concept even under demanding conditions.

\begin{figure}[htbp]
\centering
\includegraphics[width=1\linewidth]{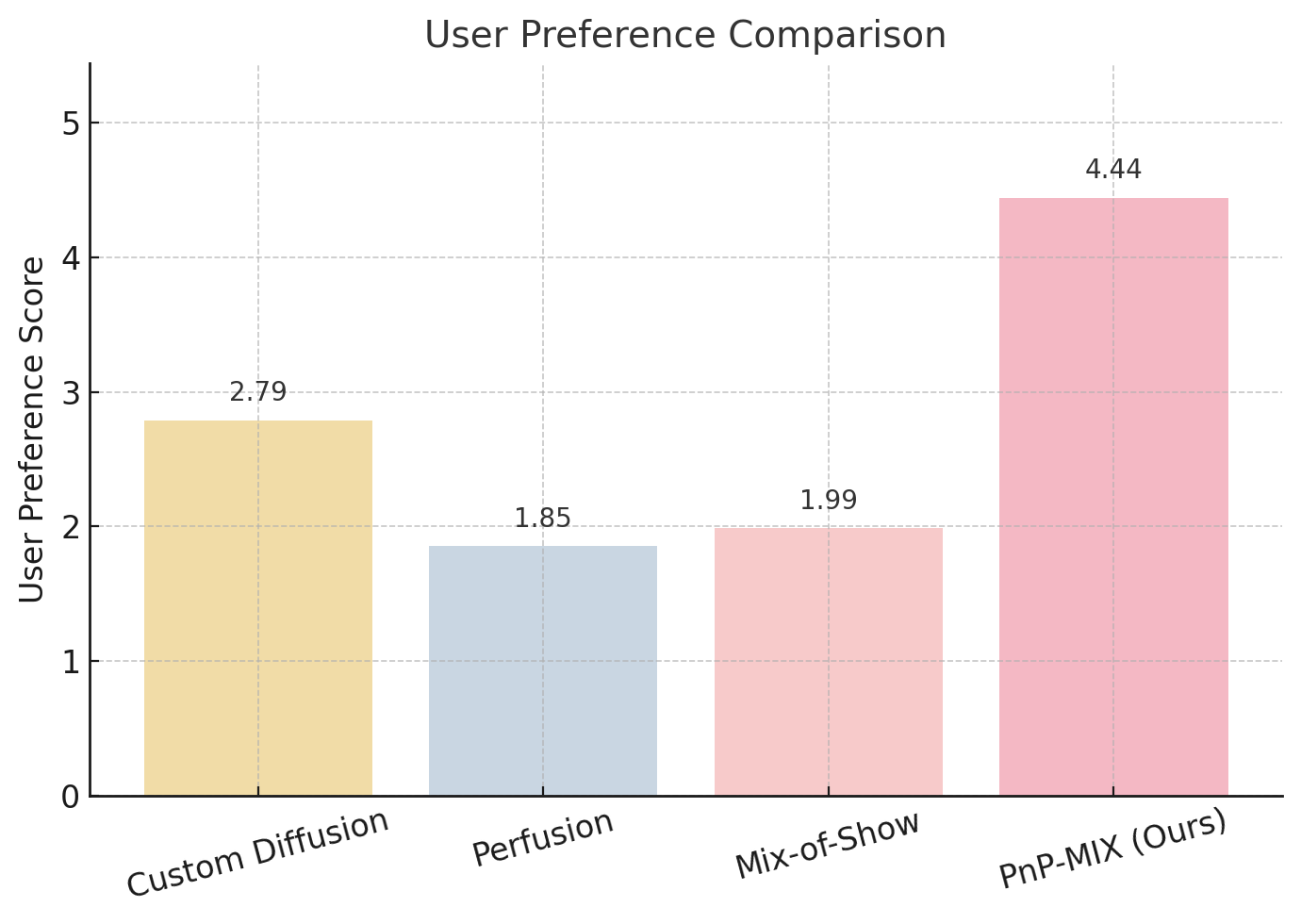}
\caption{\textbf{User preference comparison.} 
Average user ratings from a study involving 30 participants, each evaluating 32 image sets based on prompt relevance, concept accuracy, and visual realism using a 5-point scale. 
Our method (PnP-MIX, labeled as (d)) consistently outperformed baseline methods, receiving the highest preference scores across all criteria.}
\label{fig:userstudy}
\end{figure}

\textbf{User preference evaluation.} We conducted a user study involving 30 participants, each completing 32 trials~\cite{kwon2024concept}. Each trial presented participants with two reference images depicting personal concepts (e.g., a guitar and a cup), the corresponding text prompt (e.g., "A photo of a guitar placed next to a cup, bar background."), and four candidate images labeled (a) through (d). Participants evaluated each candidate image using a 5-point scale (1 = Not at all, 5 = Very well), considering three criteria jointly: text match (alignment with the prompt), concept match (accurate inclusion of specified concepts), and realism (overall visual quality and authenticity).
The mean user preference scores in Figure~\ref{fig:userstudy} show that participants most preferred our method (PnP-MIX, labeled as (d)), giving it the highest ratings on all criteria, which reflects that from the users’ perspective it delivers more satisfying visual and semantic quality than the baseline methods.

\begin{figure}[htbp]
\centering
\includegraphics[width=1\linewidth]{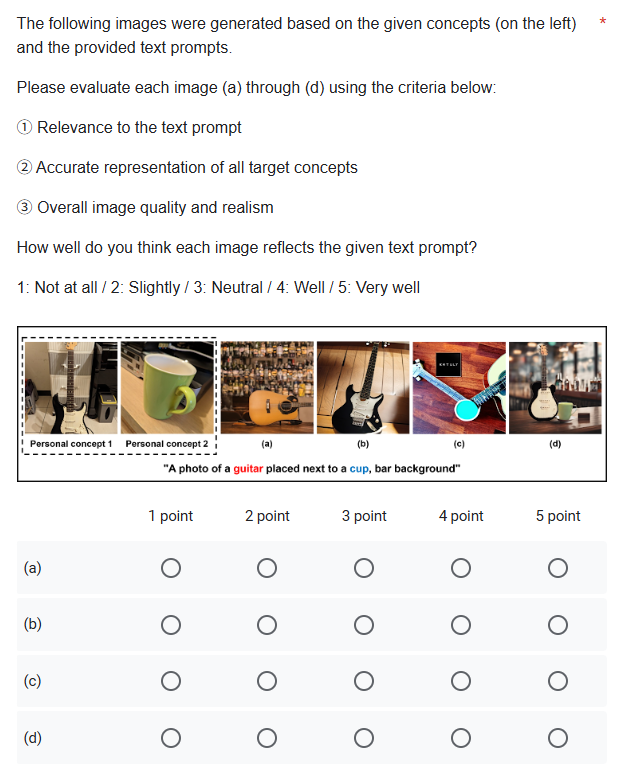}
\caption{\textbf{Questionnaire interface.} Participants saw the two reference concepts and four generated images (a–d), then rated each candidate image using a 5-point scale (1 = Not at all, 5 = Very well), considering three criteria jointly: text match, concept match, and realism.}
\label{fig:userstudysample}
\end{figure}

\begin{figure}[htb!]
\centering
\includegraphics[width=1\textwidth]{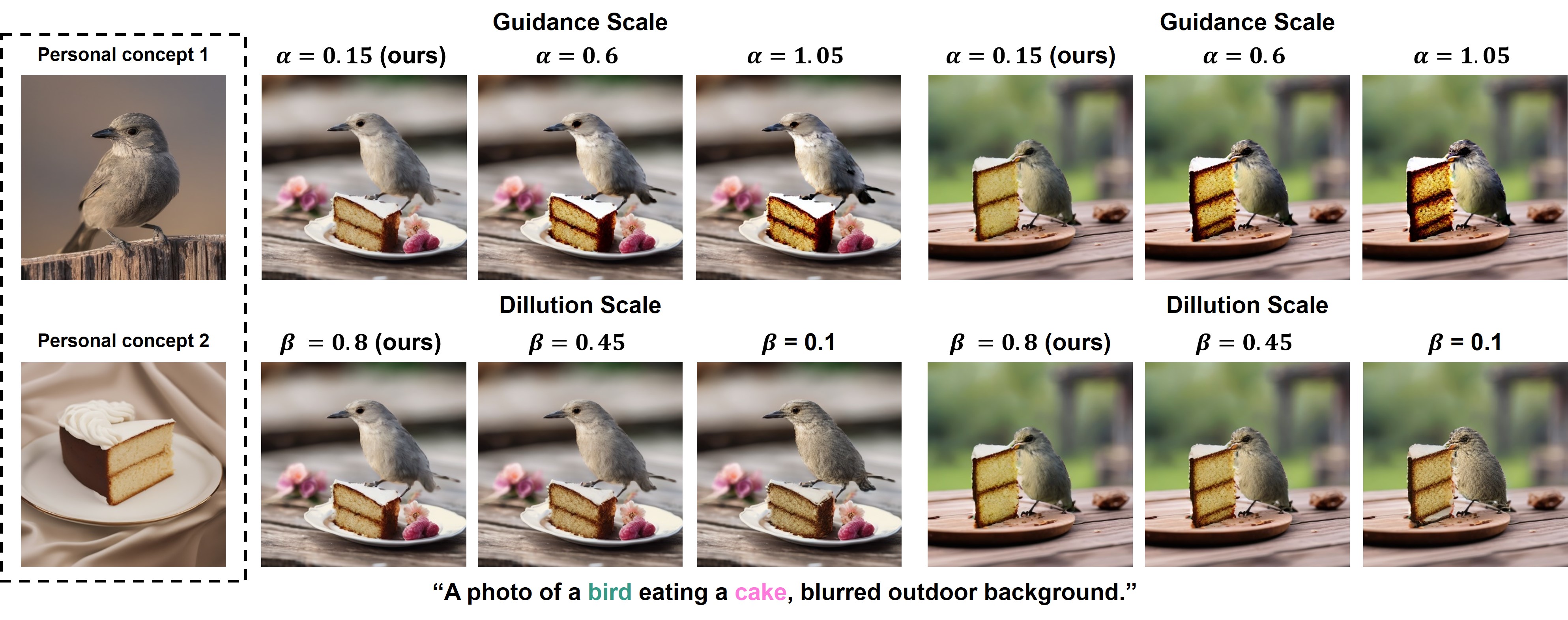}
\caption{
\textbf{Effect of guidance and dilution scales.}
Adjusting the guidance scale ($\alpha$) controls the degree of appearance transfer (e.g., color, texture) from personal concept images to target objects, while the dilution scale ($\beta$) balances the contribution of inpainted background latent and the personal concept’s appearance. Optimal hyperparameter values yield coherent multi-object generations aligned well with the prompt: \textit{"A photo of a bird eating a cake, blurred outdoor background."}}
\label{fig:hyperparameter}
\end{figure}

\textbf{Effect of hyperparameters.}
We analyzed the impact of two key hyperparameters on the compositional quality of synthesized multi-object images: the appearance guidance scale ($\alpha$) used in guided appearance attention and the background dilution scale ($\beta$) applied in background dilution++.
The guidance scale $\alpha$ modulates the extent to which appearance features from the personal concept images influence the generated objects. Higher values of $\alpha$ enhance color fidelity and texture details, resulting in outputs closely resembling their reference concepts. However, excessively large $\alpha$ values can cause oversaturated colors, creating a visual gap between the generated object and the true appearance of the personal concept. Through empirical experimentation, we determined that an $\alpha$ value of 0.15 optimally balances realistic appearance with structural coherence.
The dilution scale $\beta$ controls the extent to which the appearance of objects in the background image is diluted in the latent space.
Larger $\beta$ values substantially suppress background features near object boundaries, minimizing undesired blending and clearly delineating object edges. Conversely, when $\beta$ is too small, the original background’s appearance is insufficiently diluted, leaving unwanted background artifacts in the final image. Empirical results suggest that a $\beta$ value of 0.8 balances suppression of background features and retention of contextual coherence, effectively highlighting objects while minimizing unwanted residual background appearance.
Figure~\ref{fig:hyperparameter} illustrates these hyperparameter effects visually. Setting each parameter to an appropriate value enhances each object’s fidelity to its corresponding personal concept and improves separation from the background, thereby boosting overall image coherence and alignment with the original prompt.

\begin{table}[htbp]
\caption{\textbf{Total time comparison of multi-personalized concept generation.} PnP-MIX records the shortest overall generation time including overhead for fine-tuning or retraining, Edit-Friendly DDPM Inversion~\cite{huberman2024edit}, and inference among all evaluated methods.}
\centering
\resizebox{\linewidth}{!}{%
\begin{tabular}{clccc}
\toprule
& \textbf{Method} & \textbf{Additional training} & \textbf{Inference} & \textbf{Total} \\
\midrule
(a) & Custom Diffusion~\cite{kumari2023multi} & 814s & 13s & 827s \\
(b) & Perfusion~\cite{tewel2023key} & 7115s & 7s & 7122s \\
(c) & Mix-of-Show~\cite{gu2024mix} & 1509s & 16s & 1525s \\
(d) & \textbf{PnP-MIX (Ours)} & \textbf{0} & \textbf{151s} & \textbf{151s} \\
\bottomrule
\end{tabular}%
}
\label{tab:time_comparison}
\end{table}

\textbf{Total generation time evaluation.} Additionally, Table~\ref{tab:time_comparison} provides a comparison of computational efficiency between our method and existing approaches~\cite{ding2024freecustom}. Our reported total generation time includes all steps: additional training (fine-tuning or retraining), Edit-Friendly DDPM Inversion, and denoising inference. Since our method eliminates the need for additional fine-tuning or retraining, it significantly improves computational efficiency. In contrast, tuning-based methods entail substantial extra training costs, which limits their practical deployment. Consequently, as shown in Table~\ref{tab:time_comparison}, our method incurs zero additional training time, clearly demonstrating the advantage of our tuning-free approach. Thus, our method can be directly and efficiently applied to diverse real-world scenarios.

\begin{figure}[htb!]
\centering
\includegraphics[width=1\textwidth]{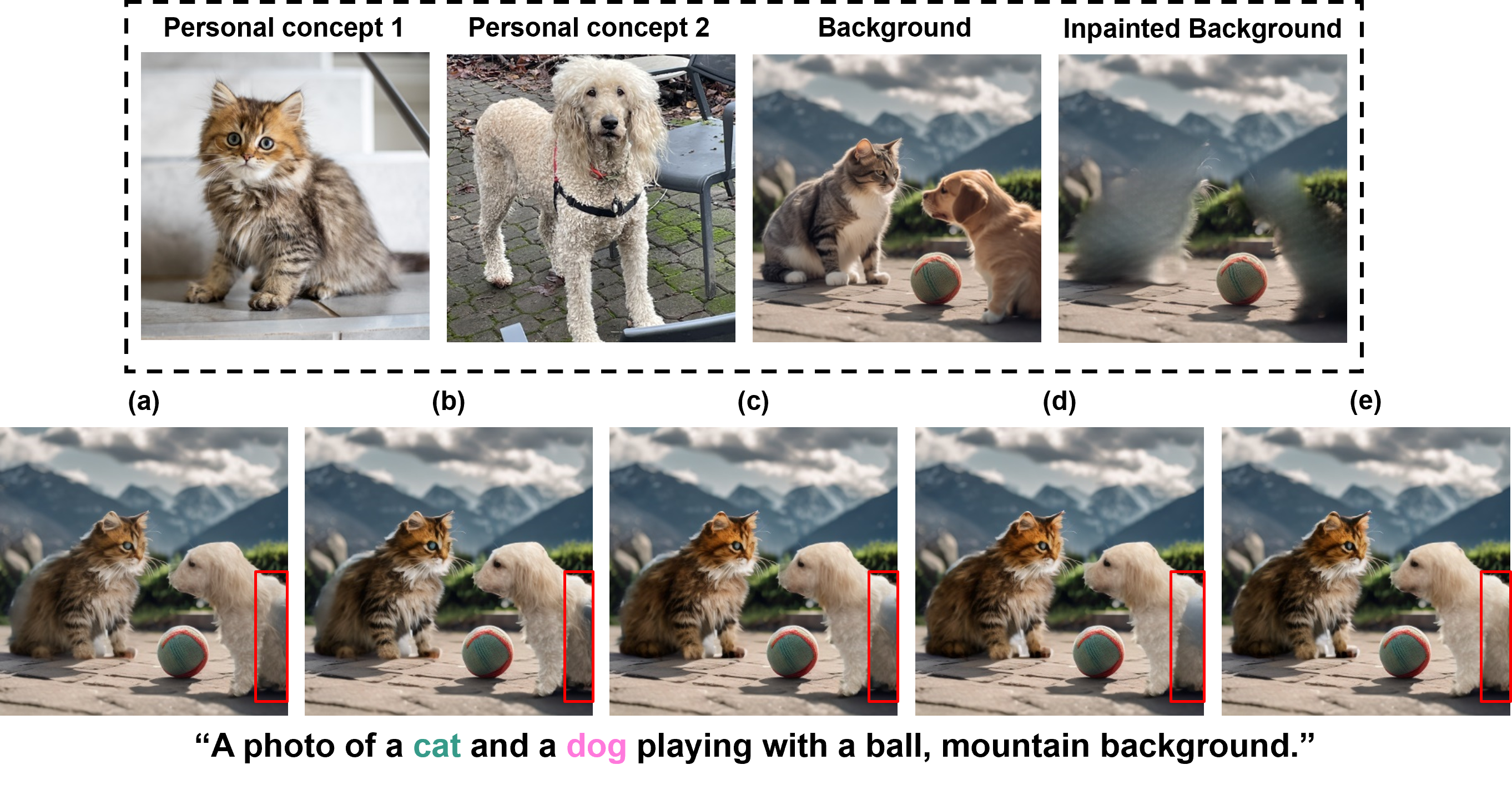}
\caption{\textbf{Ablation study.}
(a) Results using only mask-guided noise mixing and key-value replacement in self-attention. 
(b) Results from (a) with guided appearance attention. 
(c) Results from (b) with background dilution~\cite{woo2025flipconcept}. 
(d) Results from (c) with mask-guided noise mixing on the reference noise.
(e) Results from (d) with background dilution++ (\textbf{Ours}).}
\label{fig:ablation}
\end{figure}

\begin{table}[htbp]
\caption{\textbf{Ablation study.} Each component contributes meaningfully to our method's performance. The complete setting (e) achieves the best results across all metrics.}
\centering
\resizebox{1\linewidth}{!}{%
\begin{tabular}{clccc}
\toprule
& \textbf{Method} & $T_\text{CLIP} \uparrow$ & $I_\text{CLIP} \uparrow$ & $I_\text{DINO} \uparrow$\\
\midrule
(a) & Only mask-guided noise mixing            & 0.3737 & 0.8623 & 0.6005 \\
(b) & (a) + Guided appearance attention        & 0.3733 & 0.8711 & 0.6232 \\
(c) & (b) + Background dilution~\cite{woo2025flipconcept}                & 0.3743 & 0.8655 & 0.6130 \\
(d) & (c) + Mask-guided noise mixing on reference noise  & 0.3810 & 0.8830 & 0.6502 \\
(e) & \textbf{(d) + Background dilution++ (Ours)} & \textbf{0.3823} & \textbf{0.8841} & \textbf{0.6542} \\
\bottomrule
\end{tabular}%
}
\label{tab:component_comparison2}
\end{table}

\section{Ablation Study}

Table~\ref{tab:component_comparison2} and Figure~\ref{fig:ablation} jointly illustrate the contributions of each component in our framework.

Specifically, setting (a) utilizes only mask-guided noise mixing and key-value replacement~\cite{cao2023masactrl, alaluf2024cross} within self-attention layers. Although effective, it resulted in noticeable distortions and imprecise integration of target concepts, evident in blurred and overlapping areas (Figure~\ref{fig:ablation}(a)).
Setting (b) integrates guided appearance attention, significantly enhancing texture clarity, reducing previous distortions, and improving overall image coherence (Figure~\ref{fig:ablation}(b)).
In setting (c), we introduced background dilution~\cite{woo2025flipconcept}, notably improving foreground-background separation and further refining concept clarity, yet minor inconsistencies still persist (Figure~\ref{fig:ablation}(c)).
Setting (d) further applies mask-guided noise mixing directly to the reference noise, which significantly enhances the fidelity of generated images. It delivers clearer, better-defined concepts and effectively minimizes previous blending issues, as supported by quantitative metrics (Table~\ref{tab:component_comparison2}) and visual inspection (Figure~\ref{fig:ablation}(d)).
Finally, our complete configuration (e), incorporating background dilution++, effectively suppresses irrelevant details and interference from non-target regions, achieving the highest visual and textual fidelity. Notably, this final step corrects previous inaccuracies in mask alignment, clearly improving the integration of the dog's boundary on the right side, addressing issues present in setting (d) (Figure~\ref{fig:ablation}(e)). This comprehensive approach results in superior quantitative scores (Table~\ref{tab:component_comparison2}) and consistently generates images with clearly distinct and coherently integrated concepts.

Overall, our ablation study verifies that each component contributes significantly to performance improvements, confirming its essential role in successful multi-concept personalization.informative parts of the frame can help to mitigate these issues.

%% file: chapters/conclusion.tex
\chapter{Conclusion}
\sloppy

In this study, we introduced PnP-MIX, a novel method for effectively integrating multiple personalized concepts into a single image. To address issues such as undesired concept blending, background distortion, and concept leakage inherent in existing approaches, we combined unique techniques, namely guided appearance attention, mask-guided noise mixing, and background dilution++. Our method enables the generation of high-quality multi-concept images with strong visual fidelity and semantic consistency without requiring additional parameter tuning or training, making it particularly suitable for practical use.

The methodologies and insights presented here can significantly benefit future research in multi-concept personalization. The tuning-free nature of PnP-MIX offers valuable practical tools for researchers working with limited computational resources or seeking rapid development cycles. Additionally, the proposed strategies are generalizable and can be effectively utilized across various diffusion model-based image editing and synthesis applications~\cite{lee2001automatic}.

Despite its demonstrated advantages, PnP-MIX exhibits limitations in scenarios where objects are spatially overlapped, leading to potential visual distortions from rigid mask-based separations. Additionally, when the visual appearance of a personal concept significantly differs from that of objects present in the chosen background (e.g., style, texture, or scale disparities), accurately reflecting the concept becomes challenging~\cite{maeng2012nighttime}. Future work will focus on more flexible blending techniques, such as adaptive soft masking strategies, as well as appearance normalization or style adaptation modules, aiming to enhance visual quality and fidelity even in highly complex or heterogeneous scenes.